\documentclass[a4paper,conference]{IEEEtran}

\usepackage{ifpdf}
\usepackage{cite}
\usepackage[pdftex]{graphicx}
\usepackage{array}
\usepackage{mdwmath}
\usepackage{mdwtab}
\usepackage{amssymb,latexsym}
\usepackage{stfloats}
\usepackage{amsmath}
\usepackage{subfig}
\usepackage{multirow}
\usepackage{url}

\usepackage{algcompatible}

\usepackage{algorithm}

\usepackage[font={footnotesize}]{caption}

\usepackage{balance}

\usepackage[utf8]{inputenc}

\usepackage[T1]{fontenc}

\usepackage[compatible]{algpseudocode}

\renewcommand{\figurename}{Figure}

\DeclareRobustCommand*{\IEEEauthorrefmark}[1]{%
\raisebox{0pt}[0pt][0pt]{\textsuperscript{\footnotesize\ensuremath{#1}}}}

\renewcommand\IEEEkeywordsname{Keywords}

\makeatletter
\def\ps@IEEEtitlepagestyle{%
  \def\@oddfoot{\mycopyrightnotice}%
  \def\@evenfoot{}%
}
\def\mycopyrightnotice{%
  {979-8-3503-2512-6/23/\$31.00 ©2023 IEEE \hfill}
  \gdef\mycopyrightnotice{}
}

\begin{document}

\title{Benchmarking Feature Extractors for Reinforcement Learning-Based Semiconductor Defect Localization} 
\author{\IEEEauthorblockN{
Enrique Dehaerne\IEEEauthorrefmark{1,2},
Bappaditya Dey\IEEEauthorrefmark{2},
Sandip Halder\IEEEauthorrefmark{2},
Stefan De Gendt\IEEEauthorrefmark{1,2}}\\
\IEEEauthorblockA{\IEEEauthorrefmark{1}
Faculty of Science, KU Leuven, 
3001 Leuven, Belgium}
\IEEEauthorblockA{\IEEEauthorrefmark{2}
Interuniversity Microelectronics Centre (imec), 
3001 Leuven, Belgium}
{\it enrique.dehaerne@kuleuven.be}
}

\maketitle

\begin{abstract}
As semiconductor patterning dimensions shrink, more advanced Scanning Electron Microscopy (SEM) image-based defect inspection techniques are needed. Recently, many Machine Learning (ML)-based approaches have been proposed for defect localization and have shown impressive results. These methods often rely on feature extraction from a full SEM image and possibly a number of regions of interest. In this study, we propose a deep Reinforcement Learning (RL)-based approach to defect localization which iteratively extracts features from increasingly smaller regions of the input image. We compare the results of 18 agents trained with different feature extractors. We discuss the advantages and disadvantages of different feature extractors as well as the RL-based framework in general for semiconductor defect localization. 
\end{abstract}

\begin{IEEEkeywords}
Feature extraction, metrology, object detection, reinforcement learning, semiconductor device manufacture
\end{IEEEkeywords}

\makeatletter
\setlength{\footskip}{40pt} 
\def\ps@IEEEtitlepagestyle{%
  \def\@oddfoot{\mycopyrightnotice}%
  \def\@evenfoot{}%
}
\def\mycopyrightnotice{%
  \footnotesize
  \vspace*{40pt}
  \parbox{\textwidth}{%
    \centering
    \fbox{%
      \begin{minipage}{0.9\textwidth}
        © 2023 IEEE.  Personal use of this material is permitted.  Permission from IEEE must be obtained for all other uses, in any current or future media, including reprinting/republishing this material for advertising or promotional purposes, creating new collective works, for resale or redistribution to servers or lists, or reuse of any copyrighted component of this work in other works.
      \end{minipage}%
    }%
  }%
  \gdef\mycopyrightnotice{}
}

\IEEEpeerreviewmaketitle

\section{Introduction}
\label{sect:intro}
The shrinking dimensions of semiconductor patterns have resulted in increased complexity during manufacturing. This complexity has led to a need for more advanced inspection techniques to detect and analyze wafer defects. Scanning Electron Microscopy (SEM) is a valuable imaging tool capable of high resolution and high throughput. However, SEM images inherently suffer from noise. Capturing many frames to reduce noise can be time-consuming and potentially damaging to resist coatings. Consequently, researchers are increasingly focusing on Machine Learning (ML)-based SEM defect detection methods, which demonstrate superior performance on noise handling capabilities compared to traditional image processing algorithms. Moreover, ML-based approaches offer greater adaptability to variations in patterns (such as shape geometry and CD/Pitch) and imaging conditions (such as contrast and field-of-view).

We propose a deep Reinforcement Learning (RL)-based framework to automatically localize defects in SEM images. Compared to related works that don't use RL, the proposed method can iteratively inspect increasingly smaller regions of interest in the image to localize defects. Each step of this search process relies on image features of the current Region of Interest (RoI), which means that feature extraction is a critical component of an RL-based system. The two main contributions of this research work are: (i) to the best of the authors' knowledge, this is the first work to propose an RL-based semiconductor defect localization framework and (ii) we benchmark 18 different feature extractor networks for our framework and compare the best one to state-of-the-art ML-based defect detection methods.

\section{Related Work}
\label{sect:related_work}

Due to the advantages explained in the previous section, ML-based models have recently become a popular research topic for defect localization in SEM images. These include single-stage models which only extract features from an image once and predict defect locations \cite{icecs_benchmark, wang2021defectgan, dey2022yolov5}. Related to single-stage models are reconstruction-based models which extract condensed features from an input. Using these features, the model attempts to reconstruct the image and compares it to the input image to localize defects \cite{contact_hole_reconstruction,multisem_reconstruct_defect_detection}. On the other hand, multi-stage models extract features from the input image to first propose RoIs and then inspect each RoI again individually to finalize defect location predictions \cite{icecs_benchmark, semipointrend}. Some related works focus on data-centric development such as data augmentation \cite{dehaerne2023optimizing, displace_data_aug_defects} and labeling \cite{data_centric_defect_classification}.

Different from these approaches, RL-based object localization methods \cite{active_obj_loc_rl, bellver2016hierarchical, reinforcenet} iteratively inspect different input-image crops to localize objects. To the best of our knowledge, RL-based object localization methods have not yet been applied to the domain of semiconductor defect inspection. We believe RL has the potential to find defects efficiently in relatively large regions of wafers, by intelligently deciding which sub-regions to inspect more closely.

\section{Methodology}
\label{sect:methodology}

\subsection{Dataset}\label{subsect:dataset}

\begin{figure}
\includegraphics[width=\linewidth]{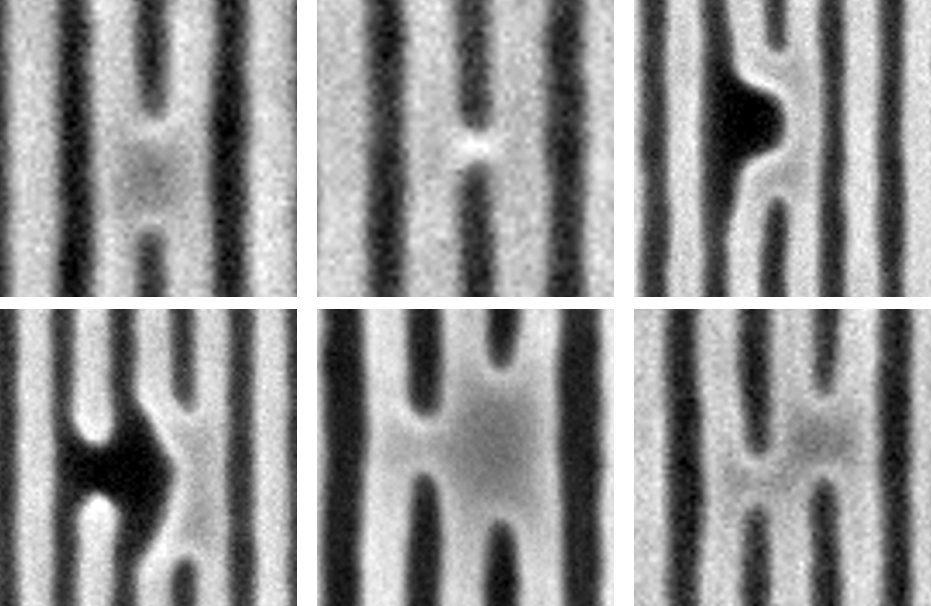} 
    \caption{Examples of each defect type in the SEM dataset. Top row (left to right): Single Bridge (SB), Thin Bridge (TB), and Line Collapse (LC). Bottom row (left to right): Line Break (LB) (along with an LC, all LBs in our dataset appear next to an LC like in this example), Multi-Bridge Horizontal (MBH), and Multi-Bridge Non-Horizontal (MBNH).}
\label{fig:defect_examples}
\end{figure}

The dataset used for our study was a collection of after-etch-inspection, line-space pattern semiconductor SEM images. Each image contained at least one defect. Each defect was categorized as belonging to one of six different defect classes. An example defect of each class is shown in Figure \ref{fig:defect_examples}. The images were split 80\% / 20\% for training/testing, respectively. Table \ref{tab:dataset_stats} shows statistics for defect instance counts and image counts of the total dataset and for each split of the dataset.

Most images contain only one defect but 103 images contained two defects. All but one of these double-defect images include at least one line collapse defect. All line break defects are located next to a line collapse defect (similar to the bottom-left example in Figure \ref{fig:defect_examples}).

\begin{table}[h]
    \caption{SEM image dataset statistics for each split.}
    \label{tab:dataset_stats}
    \begin{center}
    \begin{tabular}{|c||c|c|c|}
        \hline
        \textbf{Sample Counts} & \textbf{Train} & \textbf{Test} & \textbf{All} \\ 
        \hline \hline
        \textbf{Single Bridge (SB)} & 483 & 121 & 604\\ 
        \textbf{Thin Bridge (TB)} & 4033 & 996 & 5029\\
        \textbf{Line Collapse (LC)} & 392 & 106 & 498\\
        \textbf{Line Break (LB)} & 15 & 5 & 20\\
        \textbf{Multi-Bridge Horizontal (MBH)} & 66 & 20 & 86\\ 
        \textbf{Multi-Bridge Non-Horizontal (MBNH)} & 76 & 22 & 98\\
        \hline
        \textbf{Total defects} & 5065 & 1270 & 6335 \\
        \hline \hline
        \textbf{Total images} & 4985 & 1247 & 6232 \\
        \hline
    \end{tabular}
    \end{center}
\end{table}

\subsection{Deep RL Agent}\label{subsect:drl_agent}
\begin{figure*}
    \centering
    \includegraphics[width=\linewidth]{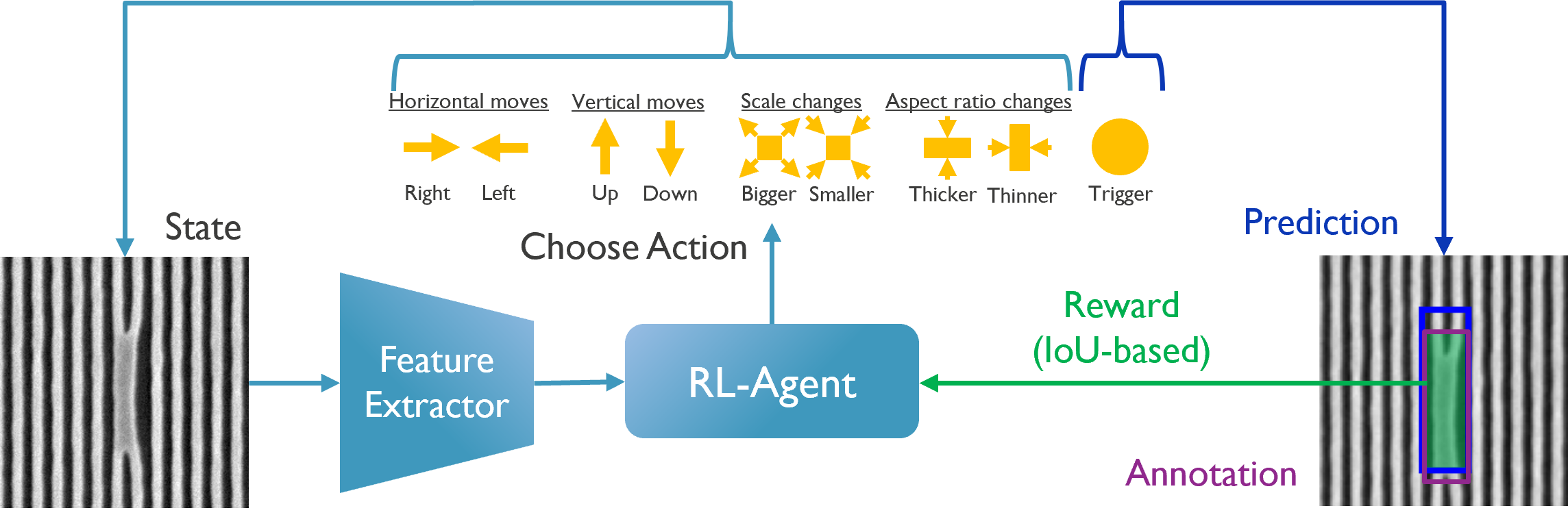}
    \caption{Overview of the proposed RL-based defect localization framework. Given an image (state), interpreted using a feature extractor, the RL agent chooses an action to modify the state. The final trigger action is chosen when the agent wants to finalize the current bounding box as its localization prediction. The IoU of the final state region and expert annotation is used as a reward signal that the RL agent uses to learn an optimal localization policy.}
    \label{fig:rl_framework}
\end{figure*}

The RL agent used in our study is similar to that of \cite{active_obj_loc_rl}. An overview of the deep RL-based defect localization system is shown in Figure \ref{fig:rl_framework}. The RL-Agent is a Deep Q-Network (DQN) \cite{deep_q_network} that chooses from 9 possible actions on the state. Each action modifies a bounding box (initially set to cover the entire input image) to localize the defect. These actions are: (i) up translation, (ii) down translation, (iii) right translation, (iv) left translation, (v) bigger scale, (vi) smaller scale, (vii) thicker aspect ratio, (viii) thinner aspect ratio, and (ix) a trigger action that finalizes the localization prediction. An action is chosen at each step based on the agent's learned policy and given inputs. 

\begin{figure*}
    \centering
    \includegraphics[width=\linewidth]{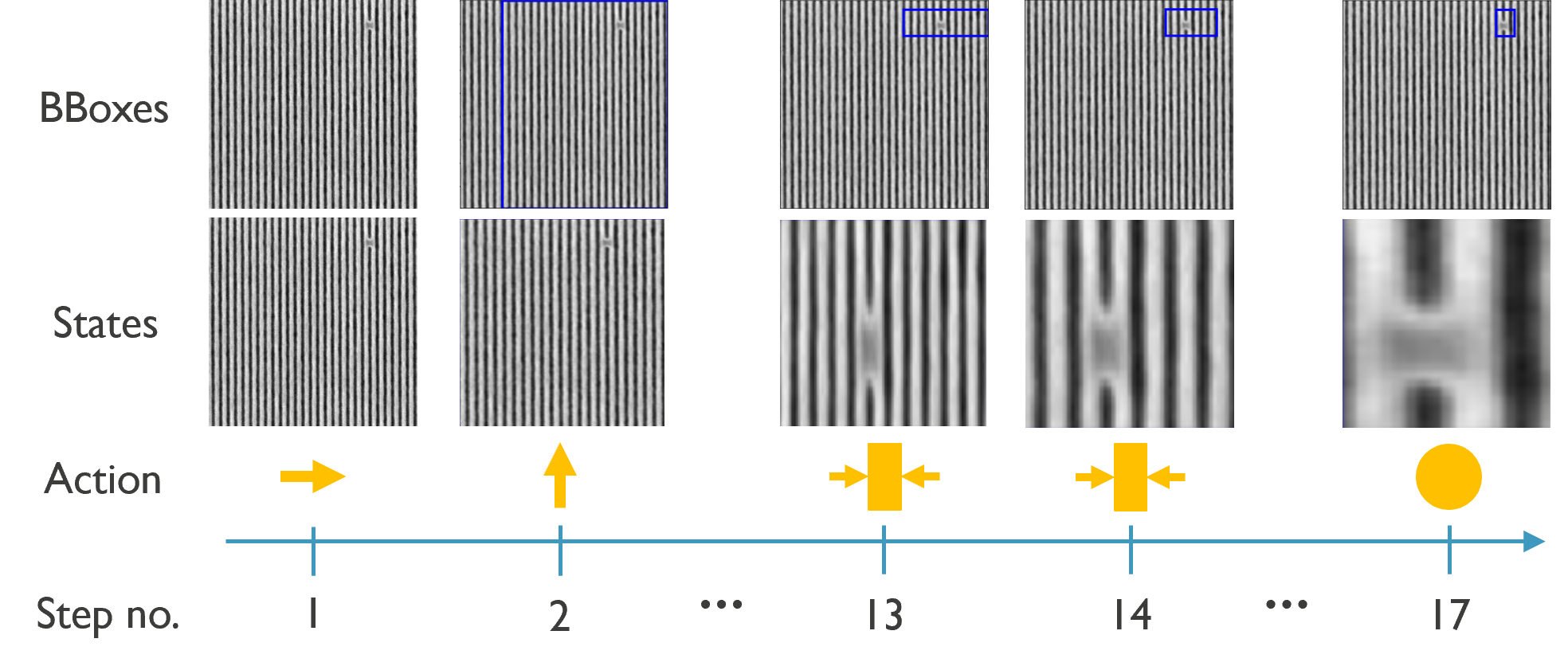}
    \caption{Sampled actions an RL agent took with corresponding states and bounding boxes to localize a single bridge defect. The states for each action are equivalent to the crop of the bounding box prediction resized to 224×224 and is given back to the RL agent (via a feature extractor) to choose the next action.}
    \label{fig:prediction_seq}
\end{figure*}

The agent takes as input a concatenation of two vectors. The first and largest vector contains the flattened, pooled features of the current state calculated by a frozen feature extractor network. The states for each subsequent step are equivalent to the crop of the image contained within the current bounding box resized to a size of 224×224. Figure \ref{fig:prediction_seq} shows an example of such a sequence of steps with the current bounding box, state, and corresponding action taken. The second vector is an encoding of the agent's action history. The RL agent is trained via a reward signal that is positive if the Intersection-over-Union (IoU) of the model's predicted bounding box and the closest annotated bounding box is above 0.5 (a commonly used IoU threshold). Otherwise, the reward is negative.

The action space of our model only allows for single-defect localization for every sequence of steps until the trigger action. To facilitate the localization of multiple objects in an image at inference time, a black cross is used as a mask to obfuscate the previously detected defect after the agent chooses the trigger action (similarly to \cite{active_obj_loc_rl}). The partially-masked image is then given back to the localization framework for another sequence of steps  to potentially localize other defects.

Our implementation is based on a publicly-available GitHub implementation\footnote{https://github.com/rayanramoul/Active-Object-Localization-Deep-Reinforcement-Learning (last cloned 27-04-2023).}. Most of the code and hyperparameters were kept the same besides three main modifications. First, the number of training episodes was set to 25 for each agent. Second, we modified the code to stop prediction on an image as soon as an agent takes 40 actions without triggering a final localization. Third, we train each agent on all defect classes. The class information was used only after testing to analyze the per-class results.

\subsection{Feature extractors}\label{sect:feature_extractors}
The feature extractor is a core component of our RL-based defect localization framework. It produces feature embeddings of the current state which is the primary input to the RL agent. The better the feature embeddings, the better the RL agent should be able to learn to localize defects. 

In this study, we investigate different feature extractor architectures, size variants, and pretraining datasets. The six architectures can be categorized as either legacy CNNs (VGG\cite{vgg16}, MobileNetV2\cite{mobilenetv2}, ResNet\cite{resnet}), recent CNNs (ConvNext\cite{convnext}), or Transformers (SwinV2\cite{swinv2} and ViT\cite{vit}). Torchvision's \cite{torchvision2016} pretrained ImageNet 1K weights were used for each model. Weights pretrained on MicroNet and Image-MicroNet (ImageNet then MicroNet) \cite{nasa_micronet_2022} were available for the legacy CNN feature extractors. MicroNet is a collection of class-annotated SEM images from various materials. We hypothesized that these weights would be better optimized for SEM images in general and therefore may lead to better feature extraction for our real fab semiconductor defect dataset.

\section{Results \& Discussion}\label{sect:results}

\begin{table*}[]
    \centering
    \caption{Test Average Precision (AP) and the average number of steps taken per image for each model trained with a different feature extractor. The names of the feature extractors correspond to their model identifiers in the torchvision model hub \cite{torchvision2016}. Numbers in \textbf{bold} indicate the best result for a given column.}
    \label{tab:test_results}
    \begin{tabular}{|c|c||c|c|c|c|c|c|c|c|}
        \hline
        \multirow{2}{*}{\textbf{Feature extractor}} & \multirow{2}{*}{\textbf{Pretrain data}} & \multicolumn{7}{c|}{\textbf{AP (@0.5 IoU)}} & \textbf{Avg no. steps}\\
        \cline{3-9}
        & & \textbf{SB} & \textbf{TB} & \textbf{LC} & \textbf{LB} & \textbf{MBH} & \textbf{MBNH} & \textbf{All (mAP)} & \textbf{(steps/image)} \\ 
        \hline\hline
        \multirow{3}{*}{\textbf{vgg16\textunderscore bn}\cite{vgg16}} & \textbf{ImageNet}\cite{imagenet} & 96.7 & 90.0 & 53.8 & 0.0 & 80.0 & 72.7 & 87.3 & 23.3 \\ 
        & \textbf{MicroNet}\cite{nasa_micronet_2022} & 95.9 & 90.3 & 51.9 & 0.0 & 80.0 & 77.3 & 87.4 & 21.4  \\
        & \textbf{Image-MicroNet}\cite{nasa_micronet_2022} & 93.4 & 93.0 & 67.9 & 0.0 & 85.0 & 77.3 & 90.7 & 21.6 \\
        \hline
        \multirow{3}{*}{\textbf{mobilenet\textunderscore v2}\cite{mobilenetv2}} & \textbf{ImageNet}\cite{imagenet} & 92.6 & 89.1 & 63.2 & 0.0 & 85.0 & 90.9 & 87.4 & 42.7 \\ 
        & \textbf{MicroNet}\cite{nasa_micronet_2022} & 87.6 & 75.0 & 58.49 & 0.0 & 85.0 & 81.8 & 75.3 & 25.0 \\
        & \textbf{Image-MicroNet}\cite{nasa_micronet_2022} & 81.0 & 63.5 & 65.1 & 0.0 & 50.0 & 50.0 & 64.9 & 27.9 \\ \hline
        \multirow{3}{*}{\textbf{resnet50}\cite{resnet}} & \textbf{ImageNet}\cite{imagenet} & 96.7 & 93.27 & 60.4 & 0.0 & 85.0 & 77.3 & 90.7 & 17.7 \\ 
        & \textbf{MicroNet}\cite{nasa_micronet_2022} & 71.1 & 59.6 & 29.2 & 0.0 & 70.0 & 72.7 & 58.7 & 23.0  \\ 
        & \textbf{Image-MicroNet}\cite{nasa_micronet_2022} & 80.2 & 53.2 & 60.4 & 0.0 & 70.0 & 86.4 & 57.4 & 40.1 \\
        \multirow{3}{*}{\textbf{resnet101}\cite{resnet}} & \textbf{ImageNet}\cite{imagenet} & 95.0 & 94.4 & 77.4 & 0.0 & \textbf{95.0} & 90.9 & 93.1 & 22.1 \\ 
        & \textbf{MicroNet}\cite{nasa_micronet_2022} & 75.2 & 61.7 & 50.9 & 0.0 & 75.0 & 68.2 & 62.6 & 53.0  \\ 
        & \textbf{Image-MicroNet}\cite{nasa_micronet_2022} & 65.3 & 44.7 & 35.8 & 0.0 & 55.0 & 50.0 & 46.2 & 16.7 \\ \hline
        \textbf{convnext\textunderscore tiny}\cite{convnext} & \multirow{2}{*}{\textbf{ImageNet}\cite{imagenet}} & \textbf{97.5} & \textbf{96.3} & 74.5 & 0.0 & 80.0 & 95.5 & 94.5 & \textbf{16.4} \\ 
        \textbf{convnext\textunderscore base}\cite{convnext} & & 96.7 & 96.0 & \textbf{80.2} & 0.0 & 90.0 & \textbf{100.0} & \textbf{94.9} & 17.9 \\ \hline
        \textbf{swin\textunderscore v2\textunderscore t}\cite{swinv2} & \multirow{2}{*}{\textbf{ImageNet}\cite{imagenet}} & 94.2 & 91.5 & 70.8 & \textbf{20.0} & 90.0 & 81.8 & 89.9 & 17.0 \\ 
        \textbf{swin\textunderscore v2\textunderscore b}\cite{swinv2} & & 95.0 & 89.3 & 73.6 & 0.0 & 85.0 & 95.5 & 88.7 & 21.4  \\ \hline
        \textbf{vit\textunderscore b\textunderscore 16}\cite{vit} &  \textbf{ImageNet}\cite{imagenet} & 81.8 & 75.2 & 67.9 & 0.0 & 60.0 & 72.7 & 75.1 & 30.2 \\ \hline
    \end{tabular}
\end{table*}

\begin{figure}
    \centering
    \includegraphics[width=0.8\linewidth]{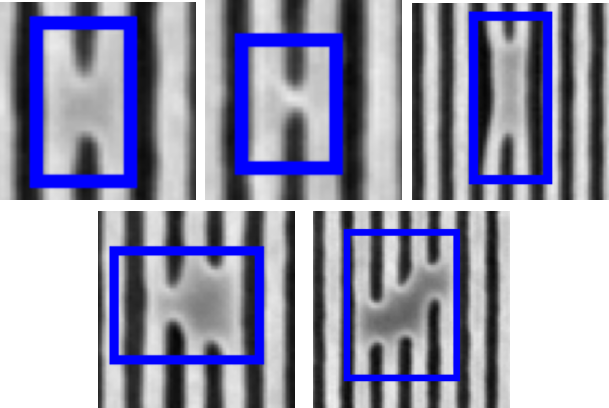}
    \caption{True positive localization examples for each of the SB, TB, LC, MBH, and MBNH defect types (from top left to bottom right, respectively) as predicted by the agent trained with the ConvNext base feature extractor \cite{convnext}.}
    \label{fig:detection_examples}
\end{figure}

Table \ref{tab:test_results} shows the test Average Precision (AP) results for each model trained with different feature extractors. The agent trained with the ConvNext base feature extractor achieved the best mean AP (mAP) of almost 94.9. Figure \ref{fig:detection_examples} shows a true-positive defect localization for each defect class (except for LB since no true positive prediction was made for LB) as predicted by this agent. The ConvNext tiny variant follows closely behind, achieving an mAP of 94.5. The agent trained with a resnet101 pretrained on ImageNet achieved the third-best mAP of 93.1. The SwinV2 models were the best transformer feature extractors. Both the base and tiny SwinV2 variants performed just under the best legacy CNNs. The ViT model did not perform well. This suggests that the coarse feature outputs of transformer models are not optimal for capturing features of small details \cite{fcbformer}, which is required for semiconductor defect inspection.

These results suggest that recent, more advanced CNN feature extractors work best for training an RL agent on the given dataset. The relative weakness of the ViT model could be because it was shown to perform optimally for very large datasets and very large model variants. In general, the sizes of annotated datasets in the semiconductor defect inspection domain are small. For this study, we only investigate the base variant of ViT to keep parameter sizes consistent with the other models. 

For all legacy CNN feature extractors, except for VGG16, the best pretraining dataset was ImageNet. The MicroNet and Image-MicroNet variants performed substantially worse. These results are unexpected since MicroNet images seem more similar to SEM defect images than ImageNet images. One possible reason for the increased performance of ImageNet pretraining is that it extracts more general features because ImageNet has far more classes than MicroNet (1000 vs 54). For VGG16, Image-MicroNet performed better than the ImageNet or MicroNet variants. This suggests that pretraining on SEM images from other domains could still be beneficial in some cases.

Along with AP, Table \ref{tab:test_results} also shows the average number of steps taken by each agent per image. Note that this metric counts steps for each predicted defect instance in an image together. The tiny ConvNext model, the second most precise model, obtains the lowest average step count of 16.4 steps/image. The base ConvNext model, the most precise model, achieves the fourth lowest step count of 17.9. This seems to suggest that model precision and average step counts are negatively correlated.  Indeed, we find a moderate negative Spearman correlation \cite{spearman_corr} of -0.50 between the mAP and the average number of steps. 

\begin{figure}
    \centering
    \includegraphics[width=\linewidth]{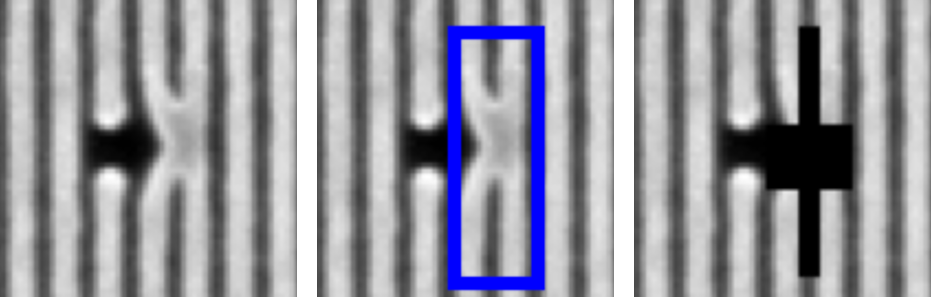}
    \caption{A failed example of a multiple-object prediction case}
    \label{fig:multiple_obj_pred}
\end{figure}

LBs were not able to be localized by almost all agents (an exception is the agent trained with tiny SwinV2 that was able to localize one of the five LBs in the test dataset). Class imbalance is a possible factor for this since LB is the least frequently occurring defect type in our dataset (15 instances in the training split). Upon further inspection, we believe another factor for the poor performance comes from the added cross used during testing to obfuscate previously detected defects. Figure \ref{fig:multiple_obj_pred} shows an example of such a scenario on a pair of LC and LB defects. The LC defect is localized first and a cross is added to hide the LC after the agent triggers it has found a defect. This cross breaks neighboring lines similar to how actual LB defects do, making it very confusing for the agent to know what is a true LB. The confusion from the black crosses and the fact that the overwhelming majority of images with two defects contain LCs could have contributed to the lower overall prediction precision for LCs compared to the other defect types.

\begin{table}
    \centering
    \caption{Bounding box mAP test results for the models from \cite{semipointrend} and our agent trained on the same dataset from \cite{semipointrend}. All feature extractors were pretrained on ImageNet.}
    \label{tab:model_comparison}
    \begin{tabular}{|c|c||c|}
    \hline
    \textbf{Model} & \textbf{Feature Extractor} & \textbf{mAP (@0.5 IoU)} \\ \hline
    \textbf{Mask R-CNN} \cite{maskrcnn_og} & \multirow{2}{*}{\textbf{resnet101 (finetuned)}} & 87.1 \\ 
    \textbf{SEMI-PointRend} \cite{pointrend_og}  & & \textbf{99.4} \\ \hline
    \textbf{DQN (Ours)} & \textbf{convnext\textunderscore base (frozen)} & 86.1  \\ \hline
    \end{tabular}
\end{table}

To compare the proposed RL-based framework to state-of-the-art models, we train and evaluate our best model on the dataset from \cite{semipointrend}. This dataset comes from the same distribution of images as the dataset described in Section \ref{subsect:dataset} and contains defects of the same classes, except for LB which are not included. Table \ref{tab:model_comparison} shows the mAP results of the two models from \cite{semipointrend} and our proposed framework using the ConvNext base feature extractor. Our framework achieves an mAP of 86.1, just under Mask R-CNN \cite{maskrcnn_og} that achieves an mAP of 87.1. However, the more advanced PointRend-based \cite{pointrend_og} model achieves the best mAP of 99.4. Note that the advantages of the Mask R-CNN and SEMI-PointRend models are that their backbones were finetuned during training while ours was not and they used data augmentation during training. An advantage of our model is that our evaluation was class-agnostic.

\section{Future Work}\label{sect:future_work}
Future work should investigate different strategies for scaling pretraining data and model sizes for feature extractors. This may include self-supervised pretraining on unannotated semiconductor SEM images. Additionally, fine-tuning performance should be compared between feature extractors pretrained on ImageNet and (Image-)MicroNet.

The results of the LB and LC defect types indicate that a better method for multiple-object prediction should be used. A straightforward solution might be replacing the black cross obfuscation artifact with a more distinct artifact. More generally applicable solutions will most likely have to modify the action space and/or action history knowledge of the RL agent.

We believe the most appropriate scenario for using the proposed RL agent is large-field-of-view SEM image inspection. This is because the agent can efficiently decide which regions of large images should be inspected more closely. Future work should validate this with large field-of-view SEM images.

\section{Conclusion}\label{sect:conclusion}
In this study, we proposed a deep Reinforcement Learning (RL)-based approach for defect localization. Compared to related works, this RL-based framework intelligently searches an SEM image step-by-step for regions that contain defects. We compared the results of 18 agents trained with different feature extractors, a critical component of the proposed localization system. We show that agents trained using ConvNext feature extractors can most precisely find defects with the lowest number of inspection steps and are competitive with previously proposed defect localization methods. We believe future work can improve our results by using more advanced pretraining/finetuning strategies and using more advanced multiple-object localization methods.

\bibliographystyle{ieeetr}
\bibliography{reference}

\end{document}